\def\year{2022}\relax
\documentclass[letterpaper]{article} 
\usepackage{aaai22}  
\usepackage{times}  
\usepackage{helvet}  
\usepackage{courier}  
\usepackage[hyphens]{url}  
\usepackage{graphicx} 
\usepackage{natbib}  
\usepackage{caption} 
\DeclareCaptionStyle{ruled}{labelfont=normalfont,labelsep=colon,strut=off} 
\usepackage{algorithm}
\usepackage{algorithmicx}
\usepackage{bm}
\usepackage{mathrsfs}
\usepackage{graphicx}
\usepackage{multirow}
\usepackage{array}
\usepackage{float}
\usepackage{caption}
\usepackage{times}
\usepackage{epsfig}
\usepackage{todonotes}
\usepackage{algpseudocode}
\usepackage{amsmath}
\usepackage{amsfonts}
\usepackage{dsfont}
 \usepackage{epstopdf}

\renewcommand{\algorithmiccomment}[1]{\bgroup\hfill\tiny~#1\egroup}

\usepackage{newfloat}
\usepackage{listings}
\floatstyle{ruled}
\newfloat{listing}{tb}{lst}{}
\floatname{listing}{Listing}
\title{Rethinking Crowdsourcing Annotation: Partial Annotation with Salient Labels for Multi-Label Image Classification}
\author{
    Jianzhe Lin, Tianze Yu, Z. Jane Wang
}
\affiliations{ The University of British Columbia
}

\usepackage{bibentry}
\begin{document}

\maketitle
\begin{abstract}
    Annotated images are required for both supervised model training and evaluation in image classification. Manually annotating images is arduous and expensive, especially for multi-labeled images. A recent trend for conducting such laboursome annotation tasks is through crowdsourcing, where images are annotated by volunteers or paid workers online (e.g., workers of Amazon Mechanical Turk) from scratch. However, the quality of crowdsourcing image annotations cannot be guaranteed, and incompleteness and incorrectness are two major concerns for crowdsourcing annotations. To address such concerns, we have a rethinking of crowdsourcing annotations: Our simple hypothesis is that if the annotators only partially annotate multi-label images with salient labels they are confident in, there will be fewer annotation errors and annotators will spend less time on uncertain labels. As a pleasant surprise, with the same annotation budget, we show a multi-label image classifier supervised by images with salient annotations can outperform models supervised by fully annotated images. Our method contributions are 2-fold: An active learning way is proposed to acquire salient labels for multi-label images; and a novel Adaptive Temperature Associated Model (ATAM) specifically using partial annotations is proposed for multi-label image classification. We conduct experiments on practical crowdsourcing data, the Open Street Map (OSM) dataset and benchmark dataset COCO 2014. When compared with state-of-the-art classification methods trained on fully annotated images, the proposed ATAM can achieve higher accuracy. The proposed idea is promising for crowdsourcing data annotation. Our code will be publicly available. 
\end{abstract}

\begin{figure}[ht]
  \includegraphics[width=\linewidth]{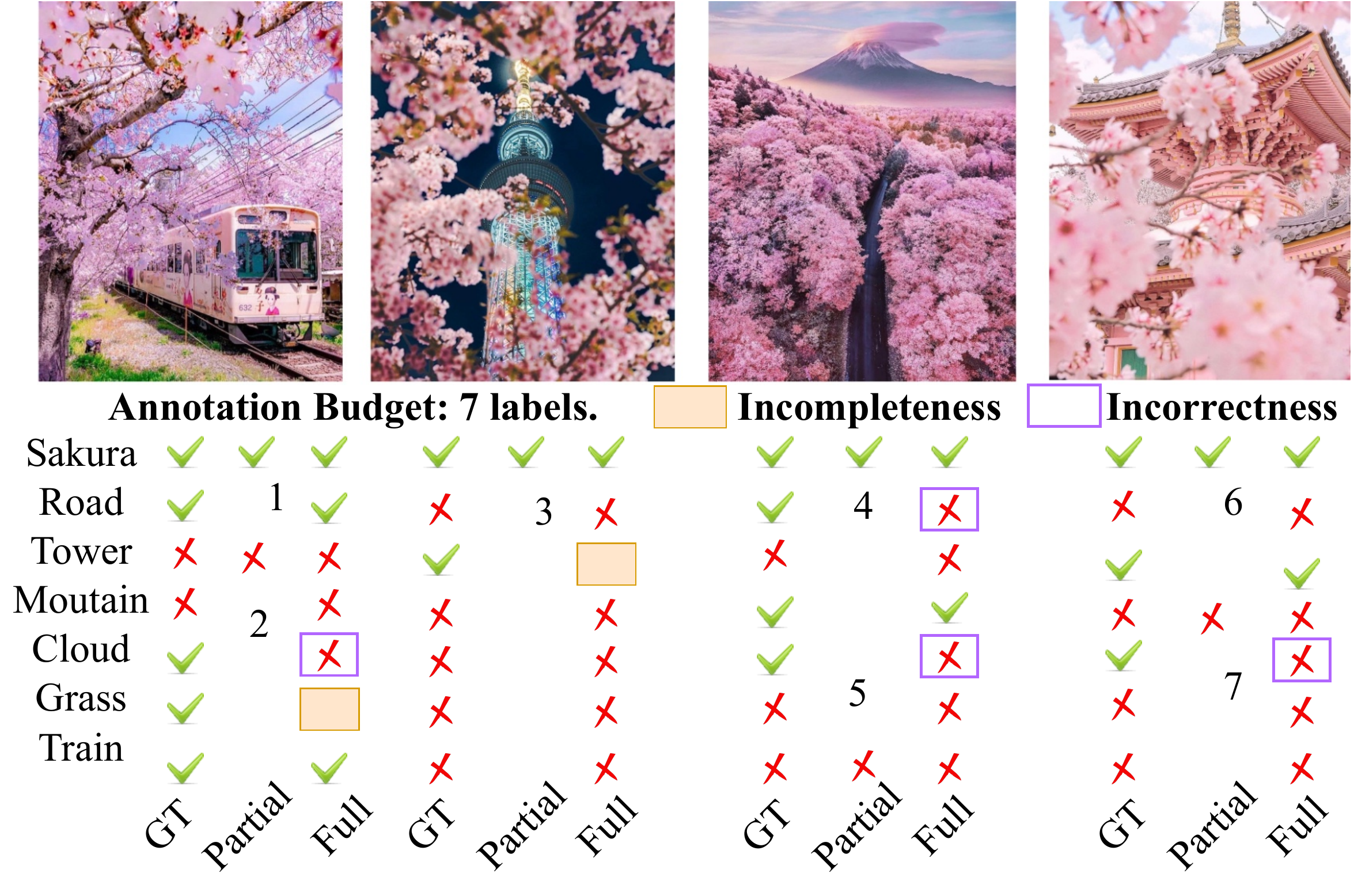}
  \caption{An illustrative comparison of partial and full crowdsourcing image annotations. GT means ground truth. Our rethinking is motivated by the following: (a) With the same annotation budget (e.g., 7 labels), either one image is fully annotated or 4 images are partially annotated efficiently/confidently with salient labels; (b) The full annotation task is more difficult. When fully annotating an image, incompleteness and incorrectness problems are common.}
  \label{fig:teaser}
\end{figure}

\maketitle

\section{Introduction}
Increasing amount of images has been generated during the last decades. Manually annotating such images is a challenging task. For example, the well-known OpenImages test set will cost $\$6.5M$ for precise annotations, and only less than 1\% OpenImages data are with annotations. The unannotated data are generally ignored when training a supervised learning model. 

Generally, supervised or semi-supervised models trained on more annotated data have better generalization abilities on new data.  Efficiently annotating unlabeled data is important, yet difficult, especially for multi-label scenes. Providing clean, multi-label annotations manually for large datasets has long been a challenging task.

A recent trend for image annotations is resorting to crowdsourcing data, such as OpenStreetMap (OSM). OSM is an editable map that is built and annotated by volunteers from scratch. However, the quality of such a manually annotated map may not be satisfying. Incompleteness and incorrectness are two primary concerns, as illustrated in the examples in Fig. \ref{fig:teaser}. Another famous crowdsourcing system is Amazon Mechanical Turk (AMT), where users (known as Taskmasters) submit their annotations in exchange for small payments. Howerver such annotations may not be reliable.

The quality of annotations could be influenced by two main factors, i.e., the annotator's reliability \cite{aydin2014crowdsourcing, demartini2012zencrowd, karger2011iterative, li2014resolving, liu2012variational} and the task's difficulty level \cite{lakshminarayanan2013inferring, li2017recover, li2020probabilistic}. 
Reliable annotators will assign the labels seriously, while unreliable ones could pick the labels carelessly in less than 10 seconds. The task's difficulty level is a major reason for label noises in crowdsourcing data. Since for a crowdsourcing platform, the annotator's reliability is hard to control, our rethinking of crowdsourcing annotation is mainly about the annotation task itself. 

In this paper, we have a rethinking of efficient and trustworthy crowdsourcing image annotation. We believe that the annotators do not need to fully annotate the images with all labels as in the traditional way. It's preferred that they partially annotate the images with the most salient labels. Annotating in this way can save the annotators the trouble of deciding uncertain labels. As they only annotate the labels they are confident in, the quality of annotations will be better and there will be fewer incompleteness/incorrectness cases. An example of partially annotating the salient labels against full annotations is shown in Fig. \ref{fig:teaser}. The major contributions of this paper are as follows: 

\begin{itemize}
    \item We propose a partial crowdsourcing annotation idea. Volunteers/workers are required to annotate the assigned annotation budget (e.g., the total number of labels), instead of annotating the assigned number of images. 
    
    \item We propose a novel active learning approach for salient label sampling. We therefore for the first time create the partially annotated datasets with salient labels.
    
    \item We propose a novel partial annotation learning algorithm, named the Adaptive Temperature Associated Model (ATAM), for multi-label image classification.
    
    \item We experimentally verify that, given the same annotation budget, the proposed classification model trained on partially annotated images outperforms the model trained with fully labeled images.
\end{itemize}

\begin{figure*}[hbt]
    \centering
  \includegraphics[width=0.8\textwidth]{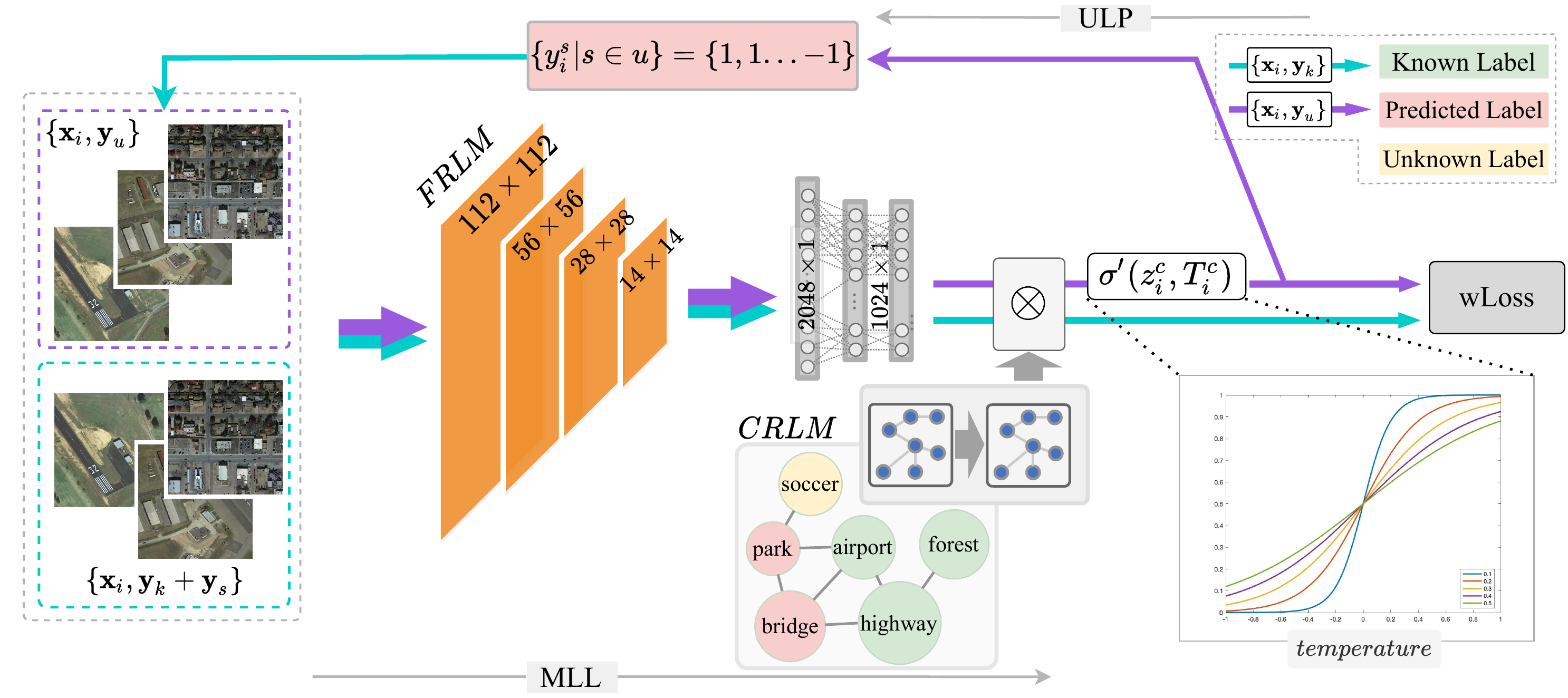}
  \caption{The training model structure of the proposed ATAM framework. The framework uses ResNet as the backbone. The ATAM mainly includes two branches, the Multi-label Learning (MLL) branch and the Unknown Label Prediction (ULP) branch. The MLL has two main modules, the feature representation learning ($\bm{FRLM}$) module and the category representation learning ($\bm{CRLM}$) module. Here wLoss is the weighted loss in Eq. (\ref{eq:wloss}), and $T$ is the smoothing factor.}
  \label{fig:framework}
\end{figure*}

\section{Related Work}
\subsection{Crowdsourcing image annotation}
Crowdsourcing services have facilitated scientists in collecting large-scale ground-truth labels \cite{aydin2014crowdsourcing, fan2015icrowd}. There are many crowdsourcing platforms, such as Amazon Mechanical Turk, which has recruited more than 500K workers from 190 countries for annotation tasks \cite{zheng2017truth, callison2009fast}. However, due to the openness of crowdsourcing, the annotators may yield low-quality annotations with noisy labels. Obtaining clean data with real ground-truth labels from the noisy labels remains a challenging topic \cite{davidson2013using, guo2012so, venetis2012max, zhang2016crowdsourced}. Such a crowdsourcing strategy might have two potential problems. First, different annotators might have different credibility and different backgrounds. For instance, an experienced annotator is more likely to provide high-quality annotations than an inexperienced annotator. Treating each annotator equally is not a wise way \cite{dawid1979maximum, demartini2012zencrowd, li2014resolving, liu2012cdas, ma2015faitcrowd}. Second, most typical crowdsourcing strategies like Majority Voting strategy are based on repeated annotation by multiple annotators to achieve higher annotation accuracy, which would be labor-expensive. In this paper, we provide an alternative way for crowdsourcing image annotations to solve the above problems: The annotators only need to partially assign the salient labels to images which they are most confident in. In this way, fewer annotators are needed for each task, and annotators are more likely to provide high-quality annotations.  

\subsection{Partial multi-label learning (PML)}
Multi-label learning has been an active research topic of practical importance, since images collected in the wild are often with more than one label \cite{tsoumakas2007multi}. The conventional multi-label learning research \cite{behpour2018arc, lyu2020partial, wang2018adaptive} mainly relies on the assumption that a small subset of images with full labels are available for training. However, such an assumption is difficult to satisfy in practice, as manually yielded annotations always suffer from incomplete and incorrect annotation problems, as illustrated in Fig. \ref{fig:teaser}. Such label concerns are especially serious for data annotated in a crowdsourcing way.  

It would be much easier if crowdsourcing annotation is only partial. With the emerging area of partial multi-label learning, it becomes possible for researchers to learn a multi-label classification model from such ambiguous data \cite{xie2020partial, sun2019partial, zhang2017disambiguation}. Traditional methods treat missing labels as negatives during the model training process \cite{mahajan2018exploring, bucak2011multi, sun2017revisiting, wang2014binary, joulin2016learning}, while it could still lead to classification performance degradation. To mitigate this problem, a novel approach for partial multi-label learning treats missing labels as a hidden variable via probabilistic models and predicts missing labels via posterior inference \cite{vasisht2014active,chu2018deep}. Differently, the work in \cite{misra2016seeing} treats missing labels as negatives, and then corrects the induced errors by learning a transformation on the output of the multi-label classifier. However, scaling the above models to large datasets is not easy \cite{deng2014scalable, huynh2020interactive}. Another recent trend for partial multi-label learning can be found in \cite{dong2018learning, liu2006semi, feng2018leveraging}, which introduces curriculum learning and bootstrapping to increase the number of annotations. During the model training, this approach uses the partially annotated data and the unannotated data whose labels the classifier is most confident with \cite{wang2019discriminative, zhang2020partial}. However, such an annotation way might introduce wrong labels that will taint the training data. This problem is called semantic drift. To better mitigate the semantic drift problem, in this paper, we propose a novel image classification model using the partially annotated data. Our motivation here is to soften the strict discrepancy between negative and positive supervisory signals by introducing a novel smoothing factor, as in Fig. \ref{fig:framework}. To achieve this goal, we simultaneously learn image and label relationships. The proposed model is named the Adaptive Temperature Associated Model (ATAM).


\section{Data Preparation}

This section employs the active learning way to prepare the salient labels for partially annotated multi-label images. 

\begin{algorithm}[t]
\scriptsize
        \caption{Active learning for annotation sampling}
        \begin{algorithmic}[1]
            \State \textbf{Given:} $X_0$ with annotation $Y_0$, the number of annotations $N_t$ in iteration $t$, and the annotation budget $L_b$.
            \State \textbf{Initialize} the pretrained network with parameter $\theta$ fine-tuned by $Y_0$.
           	\While {$L_b \geq \sum_{n=1}^t {L_n}$}
           		\State $Y_{u}^t = f(x_i;\theta)|_{i=1}^{N_t}$
           		    \State annotate labels in $Y_{u}^t$ and update the $Y_t$ with Eq. (\ref{querying}).
           		\State Get the current number of known labels $\sum_{n=1}^t {L_n}$.
           		\State Optimize $\theta$ with $Y_{t}$.
            \EndWhile
        \end{algorithmic}
        \label{active learning}
\end{algorithm}

We want to first make clear that partial annotation includes both known and unknown labels (compared with full annotations)  at the image level. While at the dataset level, the partially annotated data are used as training data.

Given a training dataset with partial annotation $\{x_i, y_{i}^{k}, y_{i}^{u}\}_{i=1}^{N}$, $y_{i}^{k}$ and $y_{i}^{u}$ denote the known labels and the unknown labels of the input image $x_i$. Note that the total number of training images $N$ is not a constant. 
Only if the volume of annotation set $Y_K$ reaches the annotation budget $L_b$, the annotation process will stop.
Suppose we have $C$ categories and $c\in C$ represents a specific category. For a deep network $f(\theta)$ ($\theta$ is the network parameters), the predicted annotation $y_{i}^{c} = f(x_i;\theta)$.
$y_{i}^{c}\in\{1, -1, 0\}$ stands for \{\textit{positive}, \textit{negative}, and \textit{unknown}\} of the category $c$ respectively. $Y_k$ and $Y_u$ represent the known label set and the unknown label set.

For the data preparation, we first annotate 50 images with salient labels, represented by $X_0$. We use these annotated samples to fine-tune the backbone network (ResNet, as in Fig. 2), which is pre-trained on ImageNet training Dataset already. Then iteratively, we input another 50 samples to the fine-tuned network and select the labels with the highest prediction confidence to annotate. We assume these labels salient labels which are also easiest to recognize to annotators. We further formulate this querying step as:
\begin{equation}
         Y_t = Y_{t-1} \cup \{ f(\bm{x}_i;\theta) \geq S\}_{i=1}^{N_t}
	\label{querying}
\end{equation}
where $S$ is the confidence of a specific class exists (for positive labels)/not exists (for negative labels) in sample $i$. We use this parameter for controlling the learning speed of active learning, as the number of labels to annotate in one iteration depends on the value of $S$. We empirically set it as 0.8. At the querying step $t$, a new subset $N_t$ (50) of samples will be actively annotated, and their annotations will be added to the current annotation set $Y_t$. The current annotated data will be further used to fine-tune the backbone network, which will be used for salient label querying in the next iteration. The query process will continue until the volume of the current annotation set $Y_t$ reaches $L_b$.
The detailed procedure is defined in Alg. \ref{active learning}. The data will be sampled and annotated iteratively in this querying-labeling way. Our goal is to generate the final annotation set $Y_k$ (equals to $Y_t$). Note that during this active learning process, we will ensure for each image at least one positive label is assigned.

\section{Adaptive Temperature Associated Model}
To tackle the problem of efficiently utilizing the annotation budget, we proposed a novel algorithm, named the Adaptive Temperature Associated Model (ATAM), for multi-label image classification using partially annotated data.
The general flowchart of the proposed model is illustrated in Fig. \ref{fig:framework}.
The framework consists of two main branches: the Multi-label Learning (MLL) branch and Unknown Label Prediction (ULP) branch.
    
\subsection{Multi-Label learning (MLL)}
For the multi-label learning branch, the goal of this branch is to optimize the framework using the annotations \{$Y_k + Y_s\}$ of the current step, where $Y_k$ is the initial known labels and $Y_s$ is the predicted labels generated by the ULP branch. 
The MLL branch consists of two modules: the feature representation learning module ($FRLM$) and the category representation learning module ($CRLM$). 
The $FRLM$ module is used to extract the features of the input image $x_i$, which is a regular CNN as in Fig. 2.
The $CRLM$ module is introduced to integrate the category information into classification.
The general operation routine of the procedure is as follows:
A $2048 \times 1$ feature vector of the input data $x_i$ is first extracted by the $FRLM$ module and fed into two fully connected(FC) layers. 
Then the output feature vector and the output of GCN(a $1024 \times C$ matrix) are fed to a scalar product layer, whose output is denoted as $z_i$. $z_i$ will be the input for the final activation layer $\sigma$, after which we will get the final predicted labels $y_i$.

\subsubsection{Category representation learning with GCN}
In multi-label classification, inner correlations between the categories contain important information.
To model the category level representation, a Graph Convolution Network (GCN) is introduced. GCN is intended to solve the problem under a non-Euclidean topological graph.
Given a graph $\mathcal{G}(V, E)$ with node features and edge features, where $V$ defines a set of nodes, $E$ is a set of edges.
For the input of the category representation learning module, GloVe\cite{pennington2014glove} is used to generate the category vectors.
The computation graph is generated based on the feature embedding of each node and its neighbors.
We follow a common practice to deploy GCN layers:
\begin{equation}
	H^{l+1} = \sigma(\hat{A}H^lW^l),
	\label{eq_GCN}
\end{equation}
where $H^l$ is the hidden representation describing the node's state at layer $l$ for the input $x_i$, $\hat A$ is the normalized correlation matrix, $W$ is a learnable transformation matrix, and $\sigma$ acts as a non-linear function where we employ LeakyReLU to implement this operation. The correlation matrix $\hat A$ here represents the co-occurence of each two labels. For example, if one sample include both label $i$ and $j$, then $A_{i,j}$ increase by 1. The final $A$ will be normalized and we will get  $\hat A$.

\subsection{Temperature Integrated ULP}
The problem we tackle in the Unknown Label Prediction (ULP) branch is to assign full annotation to the partially annotated training samples, under the limitation of the annotation budget. To be more specific, ULP branch will generate $Y_s$ (as mentioned in the above section) for the unannotated part of the training samples. 

Given the output of the last activation layer in MLL $p_i = \{p_i^k, p_i^u\}$ representing the possibilities of occurrence of known labels and unknown labels, the procedure of selecting $y_i^s$ could be formulated as:

\begin{equation}
	y^s_i = -\mathds{1} [p_i^u < (1-\beta)] \cup \mathds{1} [ p_i^u ) \geq \beta],
\label{eq:3}
\end{equation}

Here we empirically set $\beta$ as 0.8, and we start to generate $y_s$ after around 10 epochs of training. The newly generated $y_s$ in each epoch are regarded as pseudo ground truth and combined with $Y_k$ as $\{Y_k + Y_s\}$ , which will be used for the following training process.

\begin{figure}[!b]
    \centering
    \includegraphics[width=0.9\linewidth]{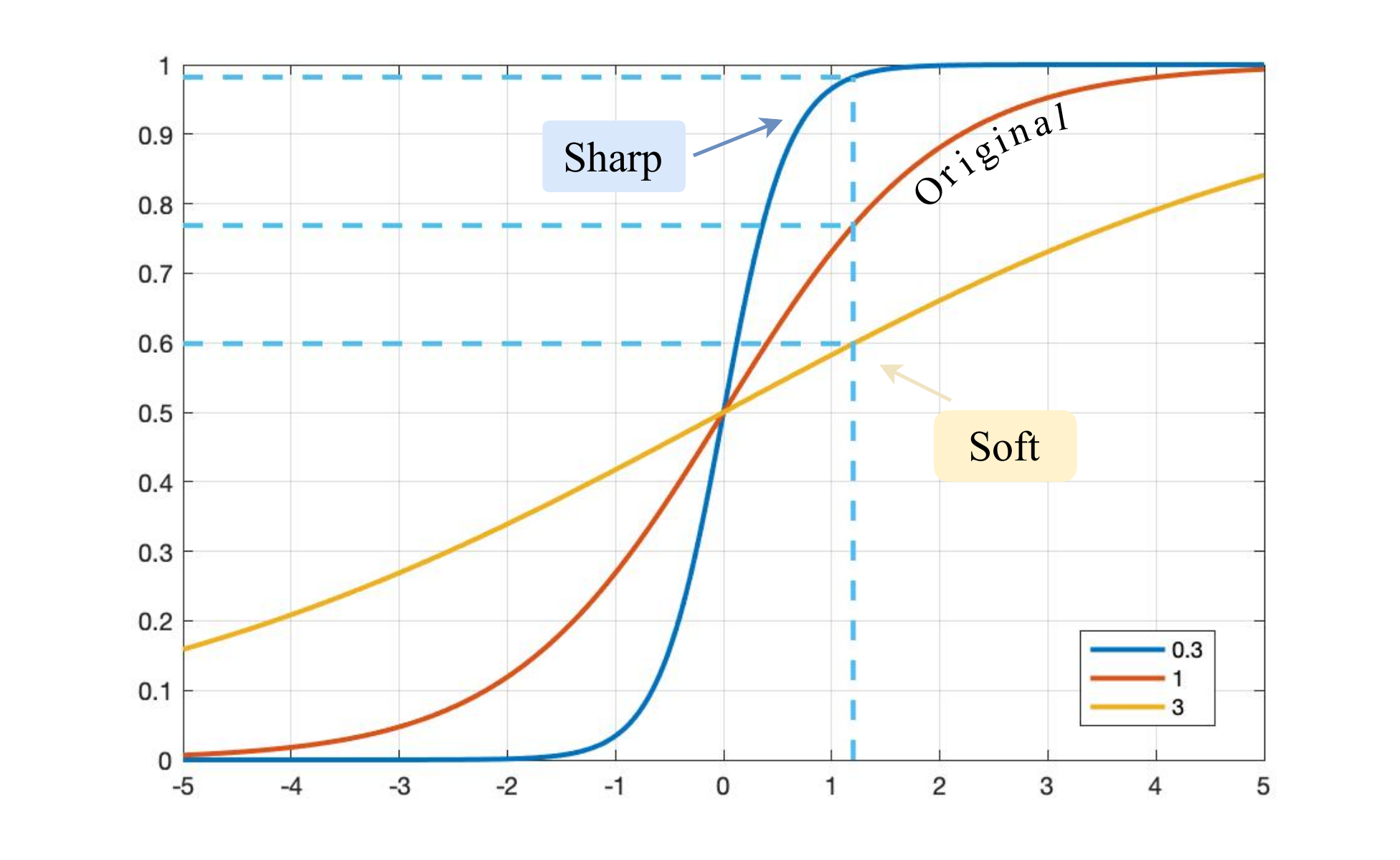}
    \caption{Illustration of how the smoothing factor $T$ works.}
    \label{fig:temperature}
\end{figure}

There exist two problems in the classification procedure. First, we need to improve the confidence level of $Y_s$ as the output $Y_s$ at the very beginning is not valid enough; Second, there must be some difficult labels whose occurrence possibilities will always stay in $[1-\beta, \beta]$, which means we cannot classify them as positive or negative. 

Inspired by the knowledge distillation idea, we propose to introduce the temperature concept to sharpen/soften the invalid/difficult label predictions. To be more specific, we introduce a temperature factor $T$ to the sigmoid function.
Integrated with the temperature factor $T$, for a random sample i and category c the sigmoid function (activation) could be defined as
\begin{equation}
	\sigma'(z_i^c, T_i^c) = \frac{1}{1+exp(-z_i^c/T_i^c)},
	\label{sigmap}
\end{equation}

z is the input of the activation layer in the neural network, which is the dot product between outputs of CRLM and FRLM. To calculate the $T$, we will introduce the edge weight in GCN as in the previous section.

We assume that an unknown label $m$ should be difficult labels if its connection $E_{m,1}, E_{m,2},...,E_{m,n}$ (suppose we have $n$ known labels) with all known labels are small, which means CRLM cannot help with the prediction. Then we will sharpen the prediction. If the connection between $E_{m,k}$ $m$ and a random known label $k$ is too large, then the prediction relies heavily on CRLM, which means the $m$ might have the same annotation with the $k$, and invalid prediction might be given at the very beginning of model training. Then we need to soften this prediction. We can also understand this assumption as we want to use the $T$ to better balance the effectiveness of CRLM and FRLM. Therefore, we define $T$ as:
\begin{equation}
	T_m^c = \alpha \cdot \mathcal{STD} (E_{mj})|_{j=1}^n ,
\end{equation}
where STD means the standard deviation and $\alpha$ is a hyper-parameter for controlling the overall smoothing level and is empirically set to be 10. A visualized function of $T$ can be found in Fig. \ref{fig:temperature}. We also want to point out that we set the maximum training epochs of ULP as 50. After 50 epochs, there are still a few difficult labels (1\%-2\%). We find the annotations of such difficult labels do not make a big difference for the following model training, and we set all these labels as negative labels. 

\subsection{Weighted focal loss and label smoothing}

An inherent problem for multi-label classification is label imbalance. There are much fewer positive samples than the negative samples. To overcome this problem, we use the weighted focal loss (wFL) instead of the regular binary cross entropy loss for the optimization of the proposed model, which is formulated as:

\begin{equation}
  \begin{aligned}
	\mathcal{L}_{k} = &- \sum\limits_{i = 1}^K {{p_\beta }} (\alpha {y_i^k}{(1 - \sigma(\cdot))^\gamma }\log \sigma(\cdot) \\ 
	&+ (1 - \alpha )(1 - {y_i^k})\sigma ^\gamma (\cdot)\log (1 - \sigma(\cdot))),
  \end{aligned}
  \label{eq:6}
\end{equation}
where $p_\beta$ is the proportion of class-wise samples with respect to all the data in the dataset and fulfills $p_\beta \in (0, 1) \& \sum_{\beta=1}^C p_{\beta} = 1$.
$\alpha$ and $\gamma$ here are empirically set as 0.25 and 2 respectively, as demonstrated in \cite{lin2017focal}.

To calculate the loss $\mathcal{L}_{s}$ for the predicted labels $Y_s$ ($Y_s$ is also taken as ground truth in the later stage of training), we introduce the smoothing factor $T$ to the activation function $\sigma$. By changing the $\sigma$ in Eq. \ref{eq:6} to $\sigma$, we can get the $\mathcal{L}_{s}$.

We further use a hyper-parameter $\epsilon$ to control the ratio of $\mathcal{L}_{k}$ and $\mathcal{L}_{s}$. we give higher priority to $Y_k$ and $\epsilon$ is empirically set as 0.5. The final loss for training is formulated as follows:
\begin{equation}
	\mathcal{L}_w = \mathcal{L}_{k} + \epsilon \mathcal{L}_{s}.
	\label{eq:wloss}
\end{equation}

\subsection{Model training}



Note that ULP branch is not trainable, as the only variable $T$ in ULP depends on the GCN part in MLL branch. The training is for the MLL branch. After getting the known label set $Y_K$ using active learning, as described in Section 3, (step-1) we firstly use the known labels $(x_i, y_i^k)$ to train the MLL branch by minimizing the loss function $\mathcal{L}_{k}$ until we get the first predicted annotations $Y_s$ (around 10 epochs).
Secondly, (step-2) with the optimized MLL branch we get the new label predictions. We use the Eq. \ref{eq:3} to get the predicted annotations $Y_s$ and extend the training annotations to $(Y_K \cup Y_s)$.
Thirdly (step-3), the MLL branch is further optimized by minimizing $\mathcal{L}_w$ using the updated training dataset $(Y_K \cup Y_s)$.
By repeating step-2 and step-3, the annotations of all categories will be acquired gradually, and the network parameter $\theta$ for the MLL branch will be optimized. After around 50 epochs, all labels for training data are acquired, and ULP branch will be removed. The following training is just regular network training for the MLL branch until convergence. For the model testing, we also only use the MLL branch to give out the multi-label prediction of each testing sample. 
A pseudo-code for model training can be found in the supplementary material.
\section{Experiments}

\begin{figure}[!b]
    \centering
    \includegraphics[width=0.7\linewidth]{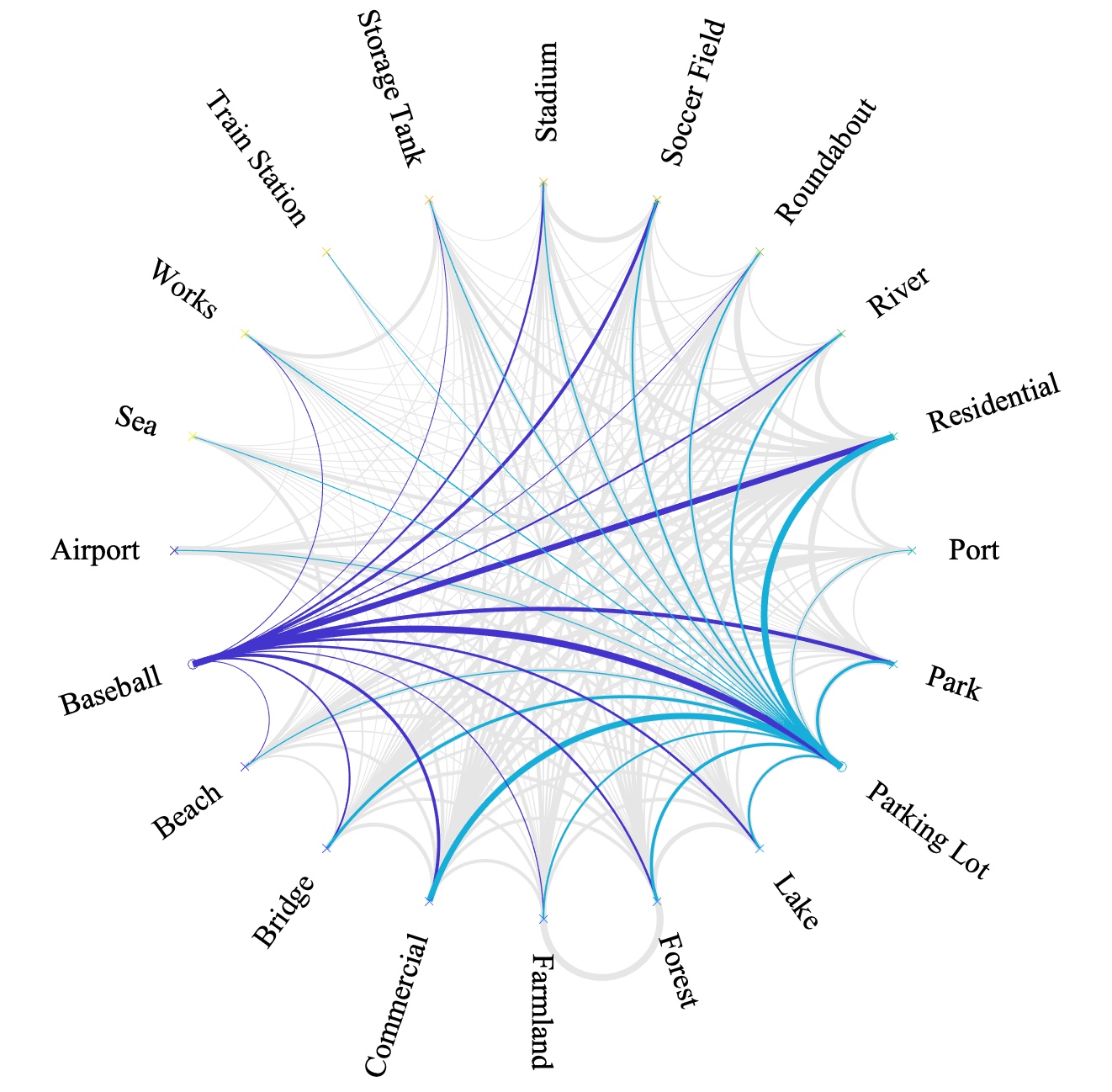}
    \caption{Label correlations of the OSM-AID dataset}
    \label{fig:osm_coore}
\end{figure}

\begin{figure}[htbp]
  \includegraphics[width=\linewidth]{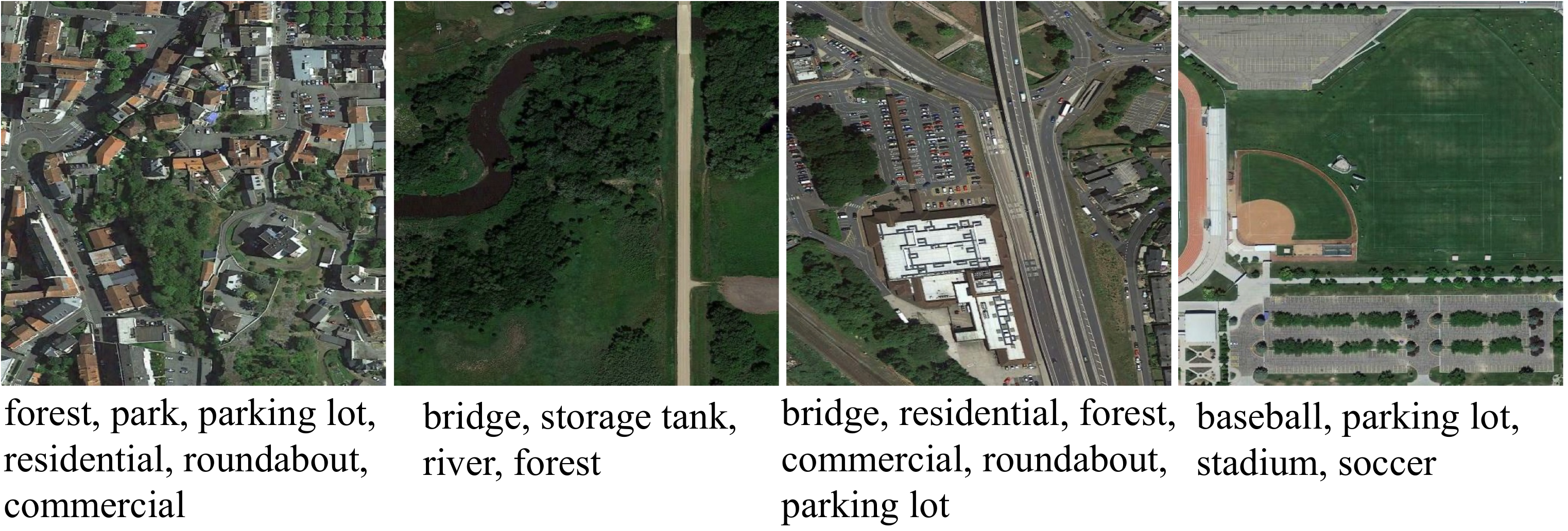}
  \caption{Example images with different number of categories from OSM-AID dataset.}
  \label{fig:OSM_exp}
\end{figure}

\subsection{Datasets, setup and evaluation metrics} 
To evaluate the performance of the proposed method, we conduct experiments on the crowdsourcing Open Street Map(OSM) dataset and traditional COCO 2014 dataset. OSM as an aerial view image data covers a large area of interest and is more difficult for classification. In our dataset, as we partially annotate the OSM with the labels from AID dataset, the created dataset is further named OSM-AID. We will release this OSM-AID data in our project website.

\begin{figure}[ht]
  \centering
  \includegraphics[width=\linewidth]{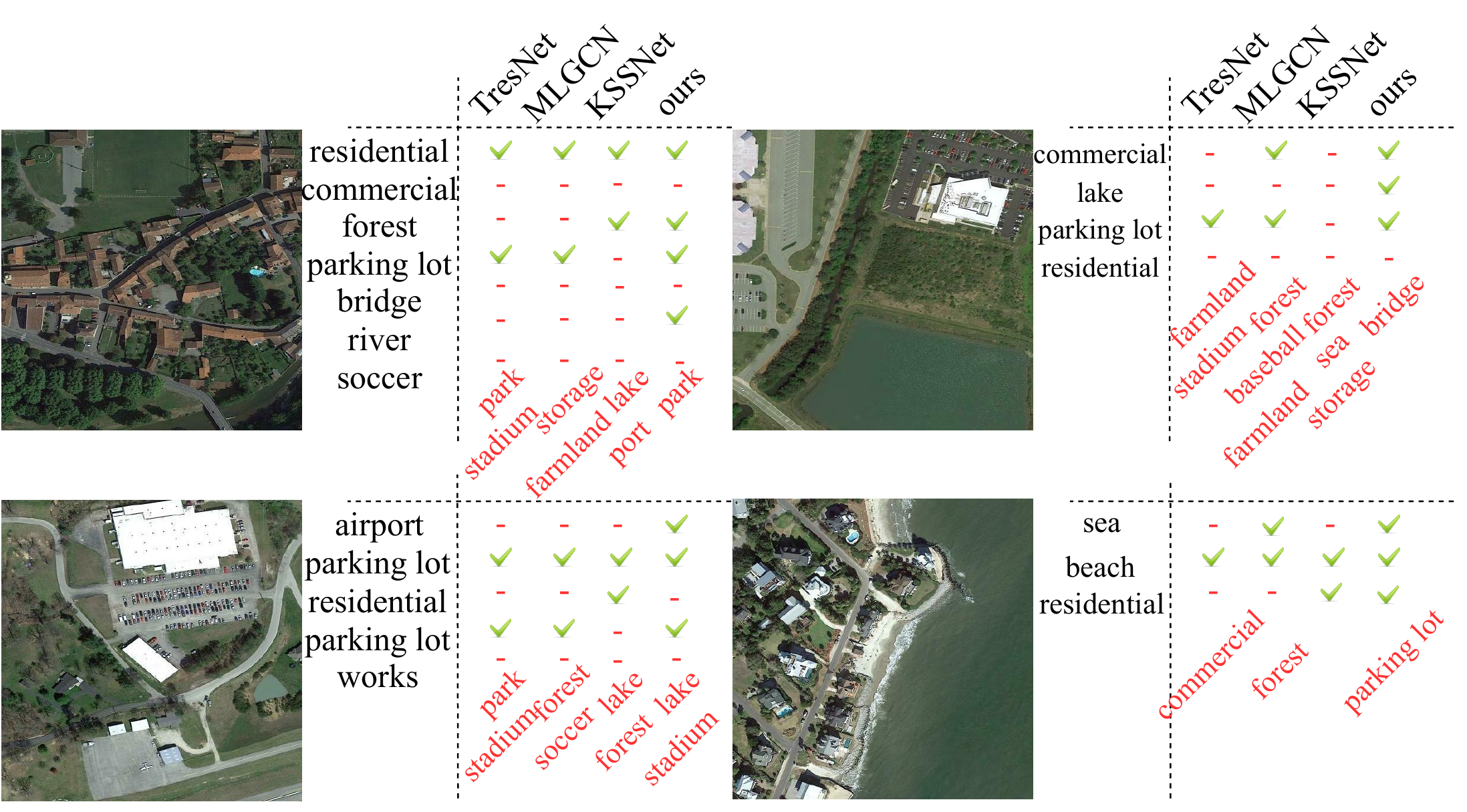}
  \caption{OSM-AID: Visualization examples of classification outputs, using TresNet (N-ML), KSSNet (N-ML), ML-GCN (N-ML) and the proposed method. Here N-ML means negative missing labels.}
\end{figure}

\textbf{OSM-AID:} Open street map with AID-label annotations has 20 classes, including Airport, Baseball, Beach, Bridge, Commercial, Farmland, Forest, Lake, Parking Lot, Park, Port, Residential, River, Roundabout, Soccer Field, Stadium, Storage Tank, Train Station, Works, and Sea.
The dataset has 3,234 images in total, and the images in each class are imbalanced, ranging from 200 to 700.
The input images are re-scaled to 512$\times$512. 
Since the OSM-AID dataset is created by ourselves, we showed examples in Fig. \ref{fig:OSM_exp}. 

To estimate the relationships between the annotations, we calculate the label correlation matrix $A$, as we mentioned in GCN.
Using labels in the training dataset, we count the times of the occurrence of different label pairs.
By visualizing the matrix $A$ of the OSM-AID dataset, as shown in Fig. \ref{fig:osm_coore}, we note that the correlations between the labels have some patterns. For example, there is `parking lot' near the `residential' or `commercial'. This co-occurrence should be considered in the modeling part, and we believe the introduction of this $A$ and the CRLM should be necessary.


Regarding the model setup, ResNet-101 \cite{resnet} is used as the backbone and is pre-trained on ImageNet \cite{imagenet}. 
For different datasets, the input images are randomly cropped and resized for data augmentation. 
SGD is used for network optimization. 
The momentum is set to be 0.9, with a decay of $10^{-4}$. 
The batch size is 12, with the input size 512$\times$512. 
The learning rate starts at 0.01 and decays by a factor of 10 for every 50 epochs.
The model is trained on Google Colab with Nvidia Tesla P100.

For performance evaluation, we report the average overall precision (OP), overall recall (OR), F1-score (OF1), F2-score (OF2). 


\subsection{Performance comparison}
In this section, we conduct several experiments to explore the best strategy to utilize the annotations with a specific budget.
For each experiment, the annotation budget is the same among different methods.

\begin{table}[ht]
\centering
\begin{tabular}{c|llll}
\hline
                  & \multicolumn{1}{c|}{} & \multicolumn{1}{c|}{} & \multicolumn{1}{c|}{} & \multicolumn{1}{c}{}  \\ [-6pt]  
                    Method         & \multicolumn{1}{c|}{OP} & \multicolumn{1}{c|}{OR} & \multicolumn{1}{c|}{OF1} & \multicolumn{1}{c}{OF2} \\ \hline \hline
                TresNet-ASL  & 0.5700 & 0.3701 & 0.4488 & 0.3980  \\
                ML-GCN 		  & 0.4620 & 0.4034 & 0.4307 & 0.4139  \\
                KSSNet 		 & 0.4704 & 0.3622  & 0.4093 & 0.3797  \\ \hline \hline
		                ATAM    & \textbf{0.6826} & \textbf{0.4567} & \textbf{0.5472} & \textbf{0.4891} \\ 
		                ATAM(40-full)    & 0.5044    & 0.3723    & 0.4284    & 0.3929    \\
                ATAM(100-full)    & 0.7331    & 0.4850    & 0.5838    & 0.5202     \\ \hline%
\end{tabular}
\caption{Classification accuracy comparisons with SOTA methods (i.e., TresNet-ASL\cite{ben2020asymmetric}, ML-GCN\cite{chen2019multi}, KSSNet \cite{wang2020multi}) and different annotation strategies. }
\label{tb:OSM}
\end{table}

\begin{table}[!htp]
\centering
\begin{tabular}{c|llll}
\hline
                  & \multicolumn{1}{c|}{} & \multicolumn{1}{c|}{} & \multicolumn{1}{c|}{} & \multicolumn{1}{c}{}  \\ [-6pt]  
Method(N-ML)            & \multicolumn{1}{c|}{OP} & \multicolumn{1}{c|}{OR} & \multicolumn{1}{c|}{OF1} & \multicolumn{1}{c}{OF2} \\ \hline \hline
                TresNet-ASL& 0.4154 & 0.2053 & 0.2748 & 0.2284\\
                ML-GCN	 & 0.6629 & 0.1277 & 0.2142 & 0.1523 \\
                KSSNet & 0.5149 & 0.1629 & 0.2475 & 0.1887  \\ \hline \hline
                ATAM    & \textbf{0.6826} & \textbf{0.4567} & \textbf{0.5472} & \textbf{0.4891} \\ \hline
\end{tabular}
\caption{Classification accuracy comparisons when incorrect negative labels exist.}
\label{tb:OSM_missing}
\end{table}



\textbf{40\% training data with full annotations vs all training data with 40\% partial annotations.}
In this experiment, three most recently proposed well-known multi-label classification algorithms, TresNet-448 \cite{ben2020asymmetric}, ML-GCN \cite{chen2019multi}, and KSSNet \cite{wang2020multi}, are used for comparison. As these SOTA methods do not have a missing label prediction module, they can only be trained with fully annotated data. For our ATAM method, we use partially annotated data for training. 
All comparison methods have the same amount of annotation budget for model training. 
To be more specific, during the implementation, we use 40\% partial annotations of training data to complete our experiments. 
We will also compare the influence of different proportions in the following sections. 

The results are shown in Table \ref{tb:OSM}. With the same annotation budget, the proposed method outperforms other multi-label methods. The reason for the better performance is that with a fixed annotation budget, the partial annotation method could access 2.5 times the number of images compared with fully annotated data.

With the training process going on, the proposed method is able to predict missing labels step by step and further use the extended annotations to optimize the model.
As can be found in Table \ref{tb:OSM}, the proposed method using partial annotations can achieve much higher accuracy than comparative methods using the fully annotated option.

\textbf{Full vs partial annotations.} We also use fully annotated data with the same annotation budget for the proposed ATAM method, to more specifically verify the effectiveness of partial annotation. 
Results can be found in Table \ref{tb:OSM}. We can find that with the same annotation budget, the partial annotation way significantly outperforms the full annotation way(ATAM(40-full)). 
Furthermore, we use a full annotation budget ($100\%$ labels) instead of $40\%$ to make a comparison with our partial method. Under such a scenario, the inputs of both methods are all the training images.
We can find that compared with training on the whole dataset with full annotations (ATAM(100-full)), our classification accuracy only slightly degrades, though it saves 60\% annotation budget.

\textbf{Simulation for data with incorrect labels.} In this section, we only use partially annotated data. We want to investigate if wrong labels exist, how accuracy changes. We assign all missing labels as negative values, the same as the crowdsourcing process, which generally takes these unannotated labels as negatives. This is also a common strategy for multi-label classification methods with missing labels. The same setting goes for both the proposed and the comparison methods. The comparison results are shown in Table \ref{tb:OSM_missing}.

For SOTA methods, compared with using less but fully annotated images as in Table \ref{tb:OSM}, the corresponding F1-score and F2-score degrade significantly. 
One main reason lies in the fact that there are quite many positive labels in the missing labels.
Assigning missing annotations with negative values will introduce too much noise to the training dataset.
Also, it exacerbates the data imbalance problem of the multi-label dataset, as we mentioned before.



\subsection{Effects of salient sampling}
In this section, we investigate the effect of active learning based on salient label sampling. In the labeling process, people always tend to provide annotations to those `easy' salient objects, especially when there is no restriction on which categories must be annotated. We want to verify the effectiveness of the salient annotation way, and we compare this way and the random annotation way in this section. Salient annotation is realized through active learning, as mentioned in the data preparation section. 
For random sampling, 40\% annotations are chosen randomly, including both positive and negative labels. For each sampled image, at least one positive label is chosen to avoid the annotated labels being all negative.
For active learning, the initial 50 images use the same strategy as the random sampling; a querying-labeling way is applied to generate the rest labels.

\begin{figure}[hbt]
  \centering
  \includegraphics[width=\linewidth]{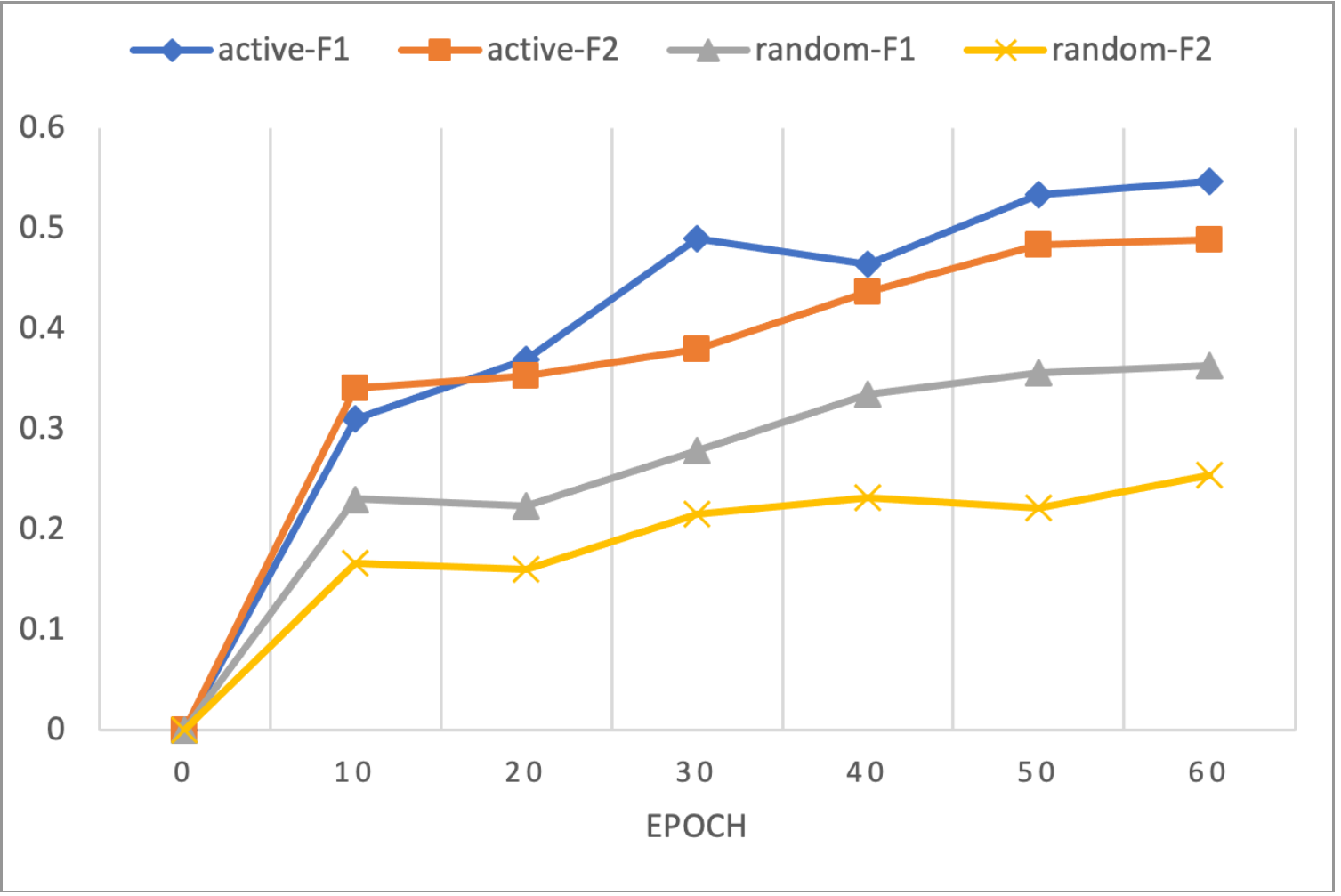}
  \caption{Random annotation versus salient annotation.}
  \label{fig:exp_active}
\end{figure}

\begin{figure}[hbt]
  \centering
  \includegraphics[width=\linewidth]{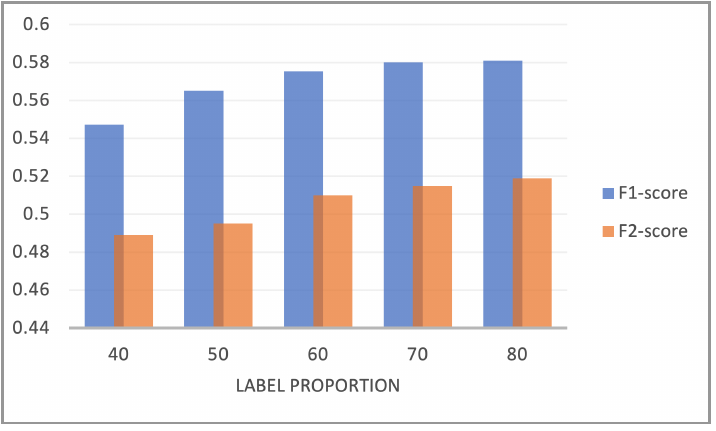}
  \caption{Performance comparisons with different label proportions.}
  \label{fig:label proportion}
\end{figure}
As shown in Fig. \ref{fig:exp_active}, salient sampling can not only accelerate the convergence speed but generates better performance compared with random sampling as well.
In other words, initializing the annotations properly could help the framework to work much more efficiently.

\subsection{Effect of label proportion}
We explore the effect of the proportion of known labels on the performance of the proposed method on the OSM-AID dataset.
As shown in Fig. \ref{fig:label proportion}, with the label proportion increase from 40 to 80 percentages, the F1-score increases from 0.5472 to 0.5801 and F2-score increases from 0.4891 to 0.5197. 
With a lower annotation budget, the proposed method could get a comparable accuracy as training with fully annotated data. 
We can conclude that annotation proportion around $50\%$ should be preferred, as the labor wasted on more annotations cannot generate a correspondingly large increase in classification accuracy.

%

\subsection{More Experiments on Benchmark Dataset}
As there's no existing benchmark partially labeled multi-label data, instead, we use the multi-label data Coco 2014. We want to further verify our proposed idea on a real scenario when there are absent labels.

\textbf{COCO2014:} is mostly used in objection detection, containing 80 categories in total, 82,081 images for training, and 40,504 images for validation and testing.
As the dataset is not specifically for multi-label classification,  the average number of categories in one image is only 3.5.
With only $40\%$ budget, it is hard to generate a partial annotation dataset fairly for different methods. 
So we randomly choose $60\%$ labels (including originally negative ones) as missing labels and set them as 'negative'. We use this way to simulate the crowdsourcing data with noise (incorrect labels). We can find the performance superiority of our proposed method in \ref{tb:COCO_missing}. This observation also shows the proposed framework is more robust when incorrect labels exist.

\begin{table}[!htp]
\centering
\begin{tabular}{c|llll}
\hline
      Method(N-ML)             & \multicolumn{1}{c|}{OP} & \multicolumn{1}{c|}{OR} & \multicolumn{1}{c|}{OF1} & \multicolumn{1}{c}{OF2}  \\ \hline \hline
                TresNet-ASL  & 0.5976 & 0.2917 & 0.3920 & 0.3250  \\
                ML-GCN		   & 0.5139 & 0.2236 & 0.3116 & 0.2521  \\
                KSSNet		 & 0.4696 & 0.2153  & 0.2952 & 0.2415  \\ \hline \hline
                ATAM    & \textbf{0.7853} & \textbf{0.5978} & \textbf{0.6788} & \textbf{0.6278}  \\ \hline

\end{tabular}
\caption{Coco 2014: Accuracy performance comparisons.}
\label{tb:COCO_missing}
\end{table}

\vspace{-3mm}
\section{Conclusion}
In this paper, inspired by our observation that partially annotating the crowdsourcing images with salient labels can decrease the annotation task difficulty and increase the annotator's reliability, we propose a novel way for crowdsourcing image annotation.  We also propose a novel framework for multi-label image classification using partially annotated images. The proposed Adaptive Temperature Associated Model can utilize the annotations more efficiently.  In the experimental part, we demonstrate that, with the same annotation budget, the classification model trained with partially annotated data can yield better performance. We will extend the proposed annotation framework to more challenging image types and learning tasks in future work.

\bibliography{aaai22}

\end{document}